\documentclass[mlabstract]{jmlr}

\usepackage{longtable}

\usepackage{booktabs}
\usepackage[load-configurations=version-1]{siunitx} 

\theorembodyfont{\upshape}
\theoremheaderfont{\scshape}
\theorempostheader{:}
\theoremsep{\newline}

\jmlrvolume{}
\firstpageno{1}
\editors{Sophia Sanborn, Christian Shewmake, Simone Azeglio, Nina Miolane}

\jmlryear{2023}
\jmlrworkshop{Symmetry and Geometry in Neural Representations}

\title{Symmetry Breaking and Equivariant Neural  Networks}

 \author{\Name{Sékou-Oumar Kaba} \Email{kabaseko@mila.quebec}\\
 \addr Mila - Quebec AI Institute\\School of Computer Science, McGill University
 \AND
 \Name{Siamak Ravanbakhsh} \Email{siamak@cs.mcgill.ca}\\
 \addr Mila - Quebec AI Institute\\School of Computer Science, McGill University
 }

\usepackage[capitalize,noabbrev]{cleveref}
\usepackage{physics}

\newcommand{\pr}[1]{\left(#1 \right)} 
\newcommand{\br}[1]{\left[#1 \right]} 
\newcommand{\cbrace}[1]{\left\{#1 \right\}} 

\renewcommand{\v}[1]{\ensuremath{\mathbf{#1}}} 
\newcommand{\gv}[1]{\ensuremath{\mbox{\boldmath$ #1 $}}} 

\DeclareMathOperator*{\argmax}{arg\,max}

\begin{document}

\maketitle

\begin{abstract}
Using symmetry as an inductive bias in deep learning has been proven to be a principled approach for sample-efficient model design. However, the relationship between symmetry and the imperative for equivariance in neural networks is not always obvious. Here, we analyze a key limitation that arises in equivariant functions: their incapacity to break symmetry at the level of individual data samples. In response, we introduce a novel notion of 'relaxed equivariance' that circumvents this limitation. We further demonstrate how to incorporate this relaxation into equivariant multilayer perceptrons (E-MLPs), offering an alternative to the noise-injection method. The relevance of symmetry breaking is then discussed in various application domains: physics, graph representation learning, combinatorial optimization and equivariant decoding.
\end{abstract}

\section{Introduction}
The notion of symmetry is of fundamental importance across the sciences, mathematics, and more recently in machine learning. It captures the idea that an object is essentially the same after some transformation is applied to it \citep{weyl1952}. Using symmetry as an inductive bias in machine learning has emerged as a powerful idea, with important conceptual and practical breakthroughs \citep{geometric_deep}.

The common intuition is that symmetry in the \textit{data distribution} should naturally lead to equivariance constraints on learned functions. However, even in symmetric domains, it appears that equivariant functions have an important limitation: the inability to break symmetry at the level of \textit{data samples}. The classical example of symmetry breaking appears in physical phase transitions. From an initially symmetric state, an asymmetric state is observed (see \cref{fig:break}). As we will see and as discussed by \citet{smidt2021}, equivariant neural networks are unable to model these phenomena. Getting rid of equivariance altogether would be an unsatisfactory solution, as it is still necessary to account for the symmetry of physical laws.

In this theory-oriented extended abstract, we give a precise characterization of this problem and argue that it is not limited to applications in physics. We show that a wide range of learning tasks require symmetry breaking and that equivariance is therefore fundamentally too constraining. We introduce a relaxation of equivariance that allows to deal with this issue. We then show how to build equivariant multilayer perceptrons \citep{symmetry_dl,ravanbakhsh2017equivariance,finzi2021practical} that can break symmetry. Finally, we propose avenues for future works and practical applications of our framework.

We introduce some mathematical background and notations used in the rest of the paper in \cref{apd:background}.

\begin{figure}

\label{fig:break}
\end{figure}






\section{Equivariance Preserves Symmetry}
\begin{figure}
\centering
\begin{subfigure}[\small{Following Curie's principle, an input cannot be mapped to an output of lower symmetry. In this example, a symmetric $0$ digit (orbit type $\br{C_4}$) cannot be mapped to a 1 (orbit type $\br{C_2}$). Likewise a 1 cannot be mapped to a 2.}][b]{
  \centering
  \includegraphics[width=.5\linewidth]{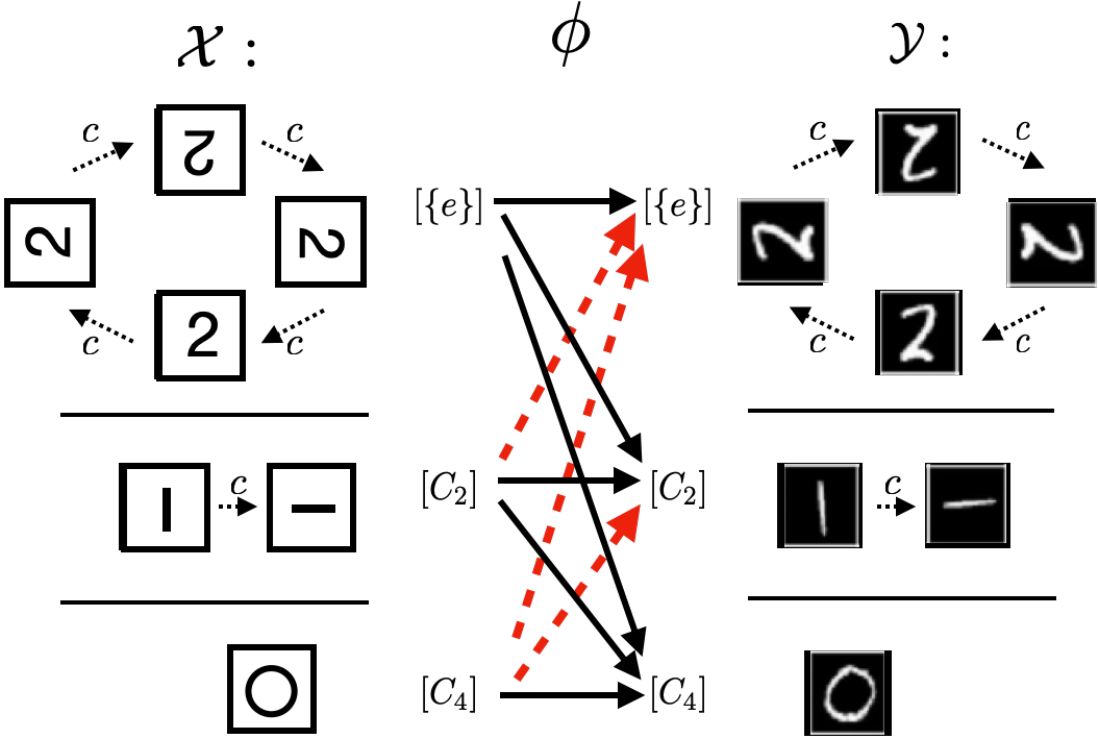}
  \label{fig:curie}
  }
\end{subfigure}%
\quad
\begin{subfigure}[\small{Relaxed equivariance solves the symmetry breaking problem by allowing any of the admissible outputs.}][b]{
  \centering
  \includegraphics[width=.3\linewidth]{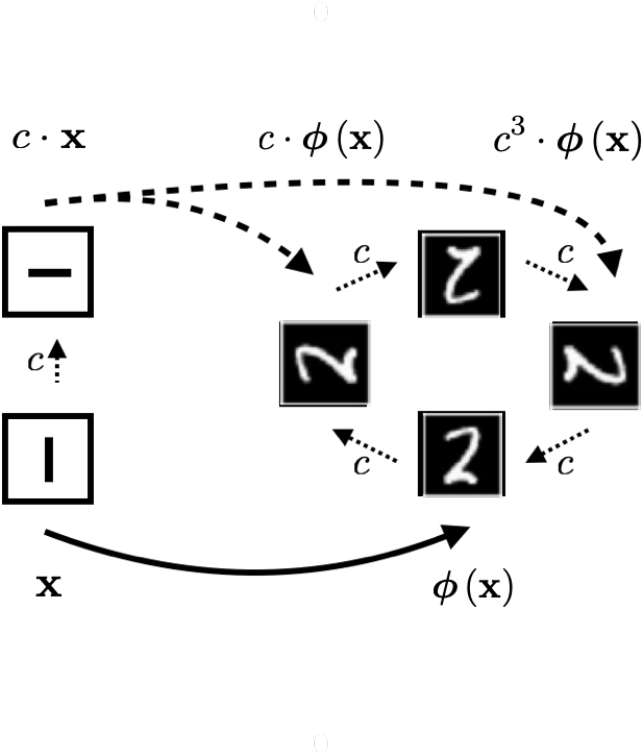}
  \label{fig:relaxed}
  }
\end{subfigure}
\caption{\small{Illustration of the symmetry breaking problem with a function equivariant to $C_4 = \langle c \rangle$. } }
\label{fig:test}
\vspace{-3ex}
\end{figure}

It is known that equivariant functions preserve the symmetry of their input. One of the earliest versions of this statement is due to \citet{curie1894symetrie}: ``\textit{the symmetries of the causes are to be found in the effects}". \citet{chalmers1970curie} provided a more mathematical version of this statement, with effects (observed state) being the result of equivariant physical laws acting on causes (initial state). The general idea is captured by the following proposition.
\begin{proposition}[Curie's Principle]
\label{th:curie}
Let $\gv{\phi}$ be an equivariant function and $G_{\v{\v{x}}}$ denote the stabilizer subgroup of $\v{x}$. Then,
\begin{align*}
G_{{\phi\pr{\v{x}}}} \geq G_{\v{\v{x}}}, \forall \v{x} \in \set{X}.
\end{align*}
\end{proposition}
The proof follows in \cref{proof:curie}.
This can also be said differently in terms of orbit types (see definition in \cref{apd:background}). When the equivariant function is seen as acting on orbits, we must have $\gv{\phi}\pr{G \cdot \v{x}} \lesssim G \cdot \v{x}$.
An equivariant function therefore cannot map an orbit of type $\br{G_{\v{x}}}$ to an orbit of type that is not coarser than $\br{G_{\v{x}}}$ (see \cref{fig:curie}).

\vspace{1ex}
For continuous functions, a version of this result holds when inputs are approximately symmetric. In this case, the inability to break the symmetry for symmetric inputs translates to more difficulty in breaking it for approximately symmetric inputs.
\begin{proposition}
\label{th:approx}
Let $\gv{\phi}$ be equivariant and Lipschitz, with constant $k$ and $\norm{\cdot}$ denote the induced norm. Then,
\begin{align*}
\norm{g\cdot \gv{\phi}\pr{\v{x}}- \gv{\phi}\pr{\v{x}}} \leq  k\norm{g\cdot \v{x} - \v{x}}, \forall g,\v{x} \in G \times \set{X}.
\end{align*}
\end{proposition}
The proof follows in \cref{proof:approx}.
If an input is close to its transformed version, the images under a continuous equivariant function also have to be close.


\vspace{1ex}
Finally, we highlight an important fact regarding symmetric inputs of finite groups.
\begin{proposition}
\label{th:mesure}
Let $\set{X} = \mathbb{R}^n$ and $\rho: G \to GL\pr{\set{X}}$ be any non-trivial linear group action of a finite group with faithful representation. Then, the set of symmetric inputs $S = \cbrace{\v{x}\in \set{X} \mid G_{\v{x}}\neq \cbrace{e}}$ is of measure zero with respect to the Lebesgue measure.
\end{proposition}
The proof follows in \cref{proof:mesure}. This captures many groups of interest in machine learning. Symmetric inputs are therefore in some sense \textit{rare}. At first glance, this could suggest that the Curie principle (Proposition \ref{th:curie}) is hardly relevant since the cases in which it would apply are improbable. Things are however not so simple. First, in many domains, such as graphs, the set of actual inputs is discrete. In this case, Proposition \ref{th:mesure} does not apply. Second, there could be a significant bias towards symmetric inputs in the data, as these data points often have special properties that make them more common. This is for example often the case in physics. Third, non-injective activation functions, like ReLUs \citep{nair2010rectified}, can make symmetric activations much more likely in the intermediary layers of a neural network by zeroing out entries. It is therefore important to handle symmetric inputs beyond the constraints imposed by equivariance, as we explain in the next section.

\section{Relaxed Equivariance}

A version of equivariance that allows breaking the symmetry of inputs and mapping to arbitrary orbit types is necessary. Some applications are detailed in \cref{sec:app}. We note that the appropriate notion was introduced by \citep{kaba2023equivariance} for canonicalization, a problem requiring symmetry breaking. However, their definition applies more generally.
\begin{definition}[Relaxed equivariance]
\label{def:relaxed}
Given group actions on $\set{X}$ and $\set{Y}$, $\gv{\phi} : \set{X}\to\set{Y}$ satisfies relaxed equivariance if $\forall g_1, \v{x} \in G \times X $, there exists $g_2 \in g_1 G_{\v{x}}$ such that 
\begin{align}
& \gv{\phi}\pr{{g_1} \cdot \v{x}} = {g_2} \cdot \gv{\phi}\pr{\v{x}}.
\end{align}
\end{definition}
The motivation for relaxed equivariance being the correct way to account for symmetry breaking is as follows. First, it captures the idea of symmetry in the task, meaning that the output of the function is predictable under transformation of the input, \textit{up to} meaningless stabilizing transformations since $\gv{\phi}\pr{{g_1} \cdot \v{x}} = {g_1} \cdot g_{\v{x}} \cdot \gv{\phi}\pr{\v{x}}$, with $g_{\v{x}}\in G_{\v{x}}$. Second, the output does not need to maintain all the symmetries of the input (see \cref{fig:relaxed}). To see this, notice that for $g_1\in G_{\v{x}}$, one possibility allowed by relaxed equivariance is $g_{\v{x}} = g_1^{-1}$. In this case, we obtain $\gv{\phi}\pr{{g_1} \cdot \v{x}} = \gv{\phi}\pr{\v{x}}$, which by contrast to what we have with equivariance (see \cref{proof:curie}), does not impose any constraints on the stabilizer of the output.

In \cref{apd:derivation}, we further justify how relaxed equivariance naturally appears in machine learning from first principles.

\section{Breaking Symmetry in Equivariant Multilayer Perceptrons}

We now investigate how to build relaxed equivariance into neural networks instead of equivariance. One seemingly ad-hoc solution is sometimes adopted to deal with symmetry breaking, for example by \cite{liu2019graph} and \cite{locatello2020object} for graph and set generation. It simply consists of adding noise to the input to break the symmetry and then using an equivariant neural network. Proposition \ref{th:mesure} confirms that this procedure has some justification. The input is almost surely mapped to a regular orbit. Then, the equivariant neural network can map the noisy input to an orbit of arbitrary type. However, there are at least two downsides to this approach. First, relaxed equivariance is only respected in expectation, similarly to equivariance when adding noise to data. Second, if the subsequent equivariant neural network is continuous, Proposition \ref{th:approx} indicates that a significant amount of noise will be required to properly break the symmetry, which might hurt generalization.

To circumvent these issues, we provide an adaptation of equivariant multilayer perceptrons (E-MLPs) that can handle symmetry breaking \citep{symmetry_dl,ravanbakhsh2017equivariance,finzi2021practical}. E-MLPs provide a standard method to build equivariant neural networks \citep{geometric_deep} and consist of stacking linear equivariant layers with point-wise non-linear functions.

Linear layers with relaxed equivariance can be constructed using the following result:
\begin{theorem}
\label{th:mlp}
Let $G$ have representations $\rho$ and $\rho'$ on $\set{X} = \mathbb{R}^n$ and $\set{Y} = \mathbb{R}^m$ respectively. Define $\set{X}_{H} = \cbrace{\v{x}\in \set{X} \mid G_\v{x} \supseteq H}$ as the invariant subspace of $\set{X}$ under $H$ and $\v{P}_{\set{X}_{H}}$ as the projection matrix onto the subspace $\set{X}_{H}$. Additionally, define $\br{H}$ be the conjugacy class of some subgroup $H \subseteq G$, and $\set{X}_{\br{H}} = \cbrace{\v{x}\in \set{X} \mid \exists K \in \br{H} s.t. G_\v{x} \supseteq K}$ to be the set of inputs stabilized by a group in $\br{H}$, e.g. inputs of type $\br{H}$.

Then, for a weight matrix $\v{W}\in \mathbb{R}^{m\times n}$, if there exists a $K\in \br{H}$ such that for all left cosets $C \in G/K$
\begin{align}
\label{eq:coset}
\pr{\v{W}- \rho'\pr{g}^T \v{W} \rho\pr{g}} \v{P}_{\set{X}_{K}} = 0,
\end{align}
where $g\in C$ is an arbitrary coset representative, then the map $\gv{\phi}: \set{X}_{\br{H}} \to \set{Y}, {\v{x}} \mapsto \v{W} \v{x}$ satisfies relaxed equivariance.
\end{theorem}
The proof and some discussion follow in \cref{proof:mlp}. Additionally, for permutation groups standard point-wise activation functions can be used, thanks to the fact that they satisfy relaxed equivariance (\cref{subsec:eq}), and that relaxed equivariance is compatible with composition (\cref{subsec:comp}).











\section{Applications}
\label{sec:app}
\begin{figure}
\centering
\begin{subfigure}[\small{Spontaneously broken symmetry in a phase transition}][b]{
  \centering
  \includegraphics[width=.37\linewidth]{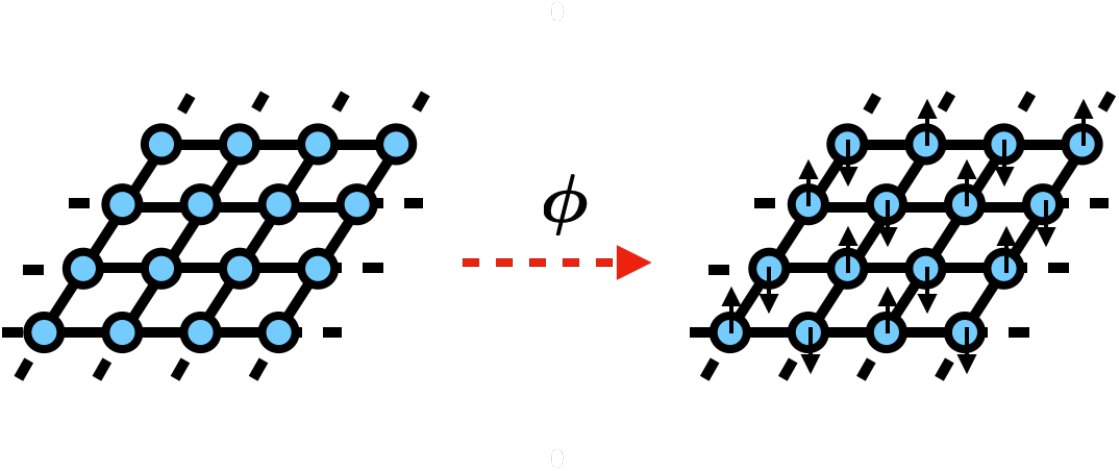}
  \label{fig:physics}
  }
\end{subfigure}%
\qquad
\begin{subfigure}[\small{Link prediction and clustering with symmetry breaking}][b]{
  \centering
  \includegraphics[width=.37\linewidth]{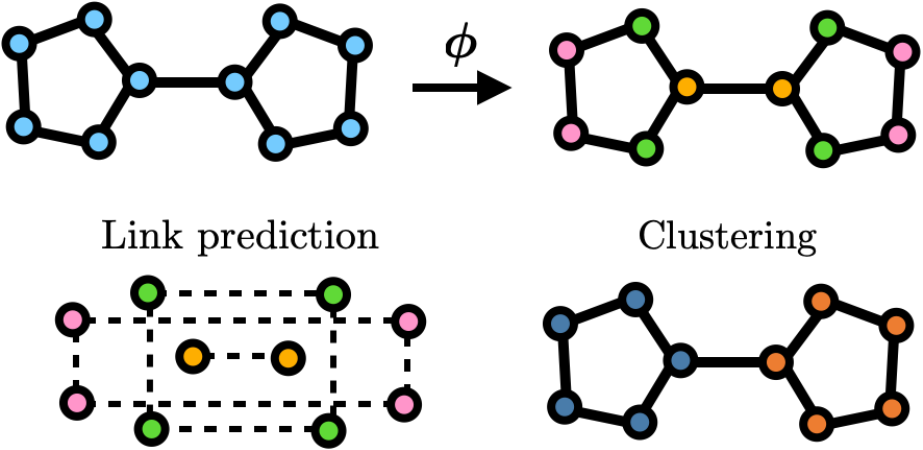}
  \label{fig:graph}
  }
\end{subfigure}
\begin{subfigure}[\small{Symmetry-degenerate solutions in a shortest path problem}][b]{
  \centering
  \includegraphics[width=.35\linewidth]{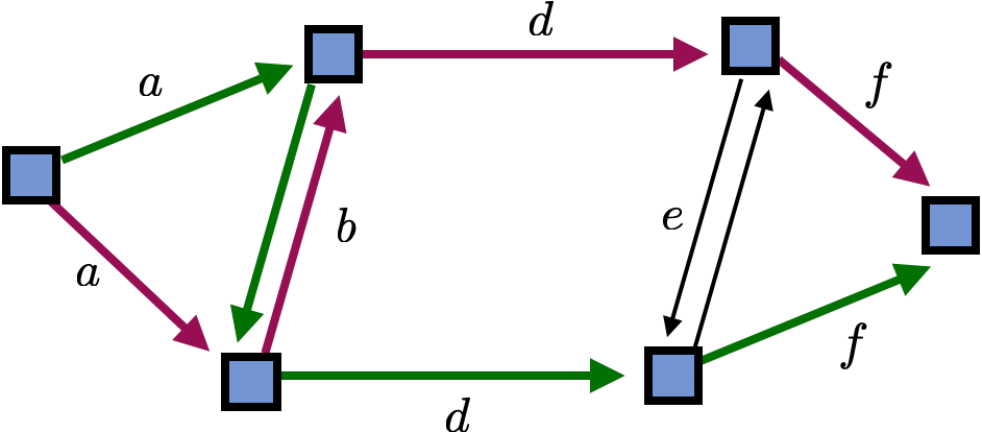}
  \label{fig:optim}
  }
\end{subfigure}%
\qquad
\begin{subfigure}[\small{Decoding from a invariant latent space requires symmetry breaking}][b]{
  \centering
  \includegraphics[width=.45\linewidth]{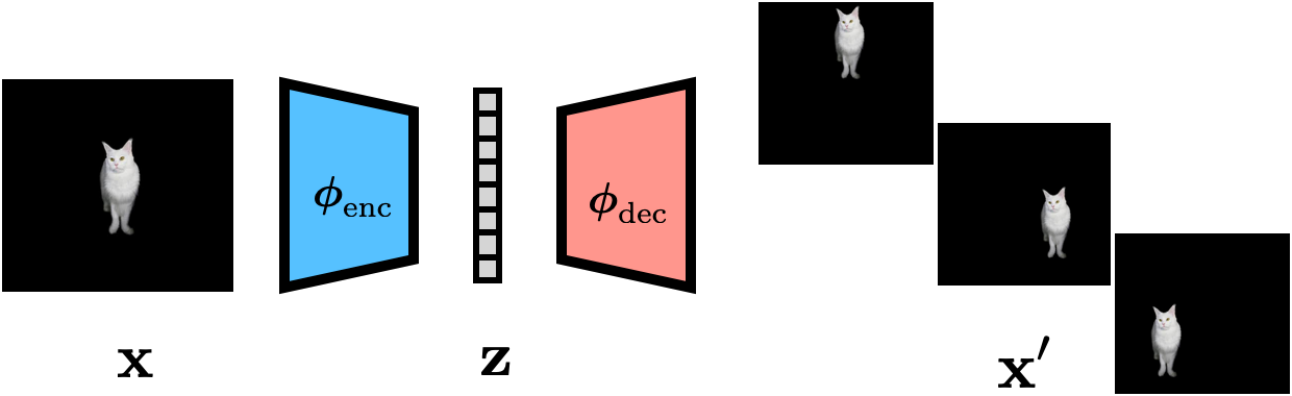}
  \label{fig:decode}
  }
\end{subfigure}%
\caption{\small{Some applications for which symmetry breaking is relevant.}}
\label{fig:examples}
\vspace{-1ex}
\end{figure}

Our analysis provides a general framework for symmetry breaking in deep learning and applies to multiple domains. We give a few examples thereafter of domains for which we think symmetry breaking analysis could be an exciting future direction (see \cref{fig:examples}).

\paragraph{Physics modelling} Symmetry breaking was first described in physics. Being able to break symmetry is important to describe phase transitions, notably in condensed matter systems \citep{kaba2022equivariant} and bifurcations in dynamical systems \citep{golubitsky2002symmetry}.
\paragraph{Graph representation learning} Learned node representations in a graph will carry the same symmetry as the graph itself. This is often not necessary and can be detrimental. The simplest example is the task of predicting edges using node representations: nodes within the same orbit of the automorphism group must be assigned the same neighbourhoods \citep{satorras2021n,lim2023expressive}.
\paragraph{Combinatorial optimization} Machine learning can be used for combinatorial optimization problems \citep{bengio2021machine}. With discrete structures, if we want to predict single solutions, it is necessary to handle degeneracies caused by symmetry -- that is we need to break the symmetry to identify a single solution.
\paragraph{Equivariant decoding} Designing a decoder from an invariant latent space to a space on which a group acts non-trivially is not a well-defined problem \citep{severo2021your,vignac2022topn, zhang2022multisetequivariant}. This can, however, be seen as an instance of symmetry breaking. This is related to the discussion of \cref{apd:derivation}.

\section{Conclusion}

In this paper, we have analyzed a fundamental limitation of equivariant functions in handling symmetry breaking. We have shown that it is important to account for it in multiple applications in machine learning by relaxing the equivariance constraint. We have finally provided a way to adapt E-MLPs to satisfy the relaxed version equivariance instead of the standard one. We hope this constitutes a first effort to better understand symmetry breaking in machine learning. Many avenues are still left to explore for the extension of this work. First, experimental testing of our claims in different domains is necessary. Second, the constraint stated in \cref{th:mlp} could be costly to solve for large groups; making it scale sublinearly in group size would be desirable. Finally, alternative ways to achieve relaxed equivariance could be explored, notably a probabilistic approach where the symmetry equivalent images are sampled instead of being deterministically computed by a network.

\acks{We wish to thank Tara Akhound-Sadegh, Yan Zhang and the reviewers for their valuable comments. This work is in part supported by the CIFAR AI chairs program and NSERC Discovery. S.-O. K.'s research is also supported by IVADO, FRQNT and the DeepMind Scholarship.}

\bibliography{biblio}

\appendix

\section{Background}
\label{apd:background}
In the following section, we introduce some useful notions on group actions and equivariant functions. The results we refer to can be found in elementary textbook on group theory, for example \cite{pinter2010book}.

\paragraph{Groups actions}
Given a group $G$ and a set $\set{X}$, a (left) group action is a function $a: G \times \set{X} \to \set{X}$, such that 
\begin{align*}
& a\pr{e, \v{x}} = \v{x} & a\pr{g, \pr{a\pr{h, \v{x}}}} = a\pr{gh, \v{x}}.
\end{align*}
We will use the shorthand notation $a\pr{g,\v{x}} = g\cdot \v{x}$ The nature of the action $\cdot$ will be clear from context.

A group action is \textit{transitive} if for any $\v{x}, \v{x}'$, there exists a $g$ such that $g\cdot \v{x} = \v{x}'$. This means that any elements can be mapped into another by the group action.

We will be mostly interested in linear group actions, for which $g \cdot \v{x} = \rho\pr{g}\v{x}$ and $\rho: G \to GL\pr{\set{X}}$ is a group homomorphism called a \textit{representation} of the group. The representation is \textit{faithful} if $\rho$ is injective.

\paragraph{Orbit types}
The \textit{orbit} of an element $\v{x}$, is defined as $G\cdot \v{x} \equiv \cbrace{g\cdot \v{x} \mid g\in G}$. It is the set of elements to which $\v{x}$ can be mapped to by the group action. The set of orbits under the group action, denoted $\set{X}/G$ forms a partition of $\set{X}$. The group action is transitive if and only if has only one orbit.

The stabilizer of an element $\v{x}$, is defined as $G_{\v{x}} \equiv \cbrace{g\in G \mid g\cdot \v{x} = \v{x}} $. $\v{x}$ is called a \textit{fixed point} or \textit{invariant} if $G_{\v{x}} = G$.
If $G_{\v{x}} = \cbrace{e}$, the action is said to be \textit{regular} on $G\cdot \v{x}$.

It can be shown that $G_{g \cdot \v{x}} = g G_{\v{x}} g^{-1} $, e.g. the stabilizers of elements in the same orbit are conjugate. We can therefore associate to each orbit a conjugacy class of a subgroup of $G$, which are the stabilizers on the orbit. Let $H$ be the stabilizer of some element of $G\cdot \v{x}$ and $\br{H} \equiv \cbrace{g H g^{-1} \mid g\in G}$ be the conjugacy class of $H$. Then, the orbit $G\cdot \v{x}$ is said to be of type $\br{H}$.

It can additionally be shown that the group action on an orbit of type $\br{H}$ is isomorphic to the group action on the cosets $G/H$ defined by 
\begin{align}
& b: G \times G/H \to G/H; & g_1, g_2H \mapsto \pr{g_1 g_2} H
\end{align}
Group actions on orbits of the same type are therefore isomorphic. We will use the symbol $\simeq$ to denote equivalence. We also introduce an order relation between orbits based on that equivalence. If $G\cdot \v{x} \simeq\br{H_1}$ and $H_1 \geq H_2$, we will say that $G\cdot \v{x} \lesssim \br{H_2}$. This is motivated by the orbit-stabilizer theorem: if an orbit has a bigger stabilizer, then it must be smaller in size.

Orbit types allow to classify group actions. Since the orbits induce a partition $\set{X}$, it is natural to decompose $\set{X}$ into orbits, with
\begin{align}
\set{X} = \bigcup_{\v{x}\in c\pr{\set{X}/G}} G\cdot \v{x} \cong \bigcup_{\v{x}\in c\pr{\set{X}/G}} \br{G_\v{x}}
\end{align}
where $c$ is a choice function over the partitions.

The kernel of a group action is defined as $\mathrm{Ker}\pr{a} \equiv \cbrace{g \in G\mid g\cdot \v{x} = \v{x}, \ \forall \ \v{x} \in \set{X}} $. It can be shown that $\mathrm{Ker}\pr{a} = \bigcap_{\v{x}\in \set{X}} G_{\v{x}} $. If an orbit is of type $\br{H}$, where $H$ is a normal subgroup, then $\mathrm{Ker}\pr{a} = H$.

\paragraph{Equivariant functions}
Given (possibly differentt) group actions on $\set{X}$ and $\set{Y}$, an equivariant function is a function $\gv{\phi}: \set{X}\to\set{Y}$ such that
\begin{align}
\gv{\phi}\pr{g\cdot \v{x}} = g\cdot \gv{\phi}\pr{\v{x}}
\end{align}
An equivariant function can therefore be seen as a homomorphism between group actions.

It follows immediately that equivariance preserves orbits
\begin{align}
\gv{\phi}\pr{G\cdot \v{x}} = G\cdot \gv{\phi}\pr{\v{x}}
\end{align}
We therefore naturally obtain that $\gv{\phi}$ induces a mapping between orbits, $\gv{\phi}: \set{X}/G\to \set{Y}/G$.

If $K$ is the kernel of the group action on $\set{Y}$, the function is considered $K$ invariant and $G/K$ equivariant. In particular, when $K=G$, the function is simply called invariant.


\section{First-principle Derivation of Relaxed Equivariance}
\label{apd:derivation}
We are in general interested in learning tasks for which the underlying distribution possesses some symmetry. For predictive modelling, given some group actions on $\set{X}$ and $\set{Y}$, that means that underlying conditional distribution satisfies $p\pr{g \cdot \v{y}|g \cdot \v{x}} = p\pr{\v{y}|\v{x}} \ \forall g\in G$. This is similar when modelling data conditioned on a latent variable with $p\pr{g \cdot \v{x}|g \cdot \v{z}} = p\pr{\v{x}|\v{z}} \ \forall g\in G$. When we wish for the model to approximate the full distribution on $\set{Y}$ (typically when $\abs{\set{Y}}$ is finite and small), equivariance with the action defined on functions follows straightforwardly. In that case, we assume ${\phi}: \set{X}\to \mathbb{P}\pr{\set{Y}}, \ \v{x} \mapsto {\phi}\br{\v{y}}\pr{\v{x}}$, where $ \mathbb{P}\pr{\set{Y}}$ is the set of probability distributions on $\set{Y}$ and obtain
\begin{align}
{\phi}\br{\v{y}}\pr{ g\cdot \v{x}} = {\phi}\br{g^{-1}\cdot\v{y}}\pr{\v{x}} = \pr{g \cdot {\phi}}\br{\v{y}}\pr{\v{x}}
\end{align}
However, in many situations, we wish to obtain a \textit{deterministic model} giving an output that maximizes the probability, rather than modelling the full distribution; e.g., Maximum a Posteriori instead of the full posterior \cite{hastie2009elements} (note that a similar argument applies when trying to approximate the distribution by a simpler one).
In this case, we define $\gv{\phi}: \set{X} \to \set{Y}$ as
\begin{align}
\gv{\phi}\pr{\v{x}} = c\pr{\argmax_{\v{y}\in \set{Y}} p\pr{\v{y}|\v{x}}}\label{eq:deterministic}
\end{align}
where the $\argmax$ is a set since the maximum may not be unique and $c: 2^{\set{Y}} \to \set{Y}$ is a choice function that selects a unique element.



We show in \cref{proof:max}, that if the distribution is symmetric under some group action, then $\argmax_{\v{y}\in \set{Y}} p\pr{\v{y}|\v{x}}$ must be a union of orbits of the stabilizer of $\v{x}$ when acting on $\set{Y}$. This is simply because, then, some probabilities are the same by symmetry.

We now assume that $\argmax_{\v{y}\in \set{Y}} p\pr{\v{y}|\v{x}}$ is a unique orbit. In a sense, this amounts to the idea that all the symmetry of the model is completely captured by the transformation group $G$. We can then prove the following theorem:
\begin{theorem}
\label{th:max}
Let $\gv{\phi}$ be defined by \cref{eq:deterministic}. If $p$ is symmetric under some action of $G$ and the set $\argmax_{\v{y}\in \set{Y}} p\pr{\v{y}|\v{x}}$ is a unique orbit, then $\gv{\phi}$ satisfies the relaxed equivariance condition. 
\end{theorem}
The proof is given in \cref{proof:max}.
Relaxed equivariance therefore naturally arises as a requirement for deterministic models under symmetric distributions. The same applies when $\gv{\phi}$ is a function that generates samples from a latent variable when the underlying conditional distribution $p\pr{\v{x}|\v{z}}$ is symmetric.

\section{Proofs}\label{apd:proofs}

\subsection{Proposition \ref{th:curie}}
\label{proof:curie}

\begin{proof}
For any $\v{x}\in \set{X}$ and $g\in G_{\v{x}}$, we have
\begin{align}
\gv{\phi}\pr{g\cdot \v{x}} = \gv{\phi}\pr{\v{x}}.
\end{align}
From equivariance of $\gv{\phi}$, we also have 
\begin{align}
\gv{\phi}\pr{g\cdot \v{x}} = g\cdot \gv{\phi}\pr{\v{x}}.
\end{align}
Thus,
\begin{align}
g\cdot \gv{\phi}\pr{\v{x}} = \gv{\phi}\pr{\v{x}}
\end{align}
The stabilizer of $\gv{\phi}\pr{\v{x}}$ is therefore at least $G_{\v{x}}$, which completes the proof.
\end{proof}

\subsection{Proposition \ref{th:approx}}
\label{proof:approx}
\begin{proof}
If $\gv{\phi}$ is Lipschitz with constant $k$, we have
\begin{align}
\norm{\gv{\phi}\pr{g\cdot \v{x}}- \gv{\phi}\pr{\v{x}}} \leq  k\norm{g\cdot \v{x} - \v{x}}, \forall g,\v{x} \in G \times \set{X}.
\end{align}
From equivariance of $\gv{\phi}$, we find 
\begin{align}
\norm{g\cdot \gv{\phi}\pr{\v{x}}- \gv{\phi}\pr{\v{x}}} \leq  k\norm{g\cdot \v{x} - \v{x}}, \forall g,\v{x} \in G \times \set{X}.
\end{align}
which completes the proof.
\end{proof}

\subsection{Proposition \ref{th:mesure}}
\label{proof:mesure}
\begin{proof}




The set $S$ is equal to $\bigcup_{g\in G/\cbrace{e}} \cbrace{x\in X\mid g\cdot x = x}$. We will show that for each $g\in G/\cbrace{e}$, the set of elements of $\set{X}$ stabilized by $g$ is of measure zero. Since the union is over a finite set, $S$ will therefore also be of measure zero.

The set of elements stabilized by $g$ is given by the solutions of the equation $\rho\pr{g}\v{x} = \v{x}$. The stabilizer is therefore the eigenspace of $\rho\pr{g}$ with eigenvalues 1. If $\rho$ is a faithful representation, then for any $g\neq e$, $\rho\pr{g}\neq I$. However, for any linear operator other than $I$, the dimension of eigenspaces with eigenvalue 1, if they exist, must be $d<n$. But, any subspace of $\mathbb{R}^n$ of dimension $d<n$ has measure zero with respect to the Lebesgue measure. Therefore, the set of elements stabilized by any $g\neq e$ is of measure zero.

This completes the proof.
\end{proof}

\subsection{\cref{th:max}}
\label{proof:max}
We introduce the following lemmas
\begin{lemma}
\label{lemma:1}
Let $ p\pr{\v{y}\mid \v{x}} = p\pr{g\cdot \v{y}\mid g\cdot \v{x}} $ for all $g\in G$. Then,
\begin{align}
G_{\v{x}} \cdot \pr{\argmax_{\v{y}\in \set{Y}} p\pr{\v{y}|\v{x}}} = \argmax_{\v{y}\in \set{Y}} p\pr{\v{y}|\v{x}}
\end{align}
\end{lemma}
\begin{proof}
From the symmetry of $p$, and based on the definition of the stabilizer, we have for all $g_{\v{x}}\in G_{\v{x}}$
\begin{align}
p\pr{\v{y}\mid \v{x}} &= p\pr{g_{\v{x}}\cdot \v{y}\mid g_{\v{x}}\cdot \v{x}} \\
p\pr{\v{y}\mid \v{x}} &= p\pr{g_{\v{x}}\cdot \v{y}\mid \v{x}} 
\end{align}
Therefore,
\begin{align}
\v{y}^* \in \argmax_{\v{y}\in \set{Y}} p\pr{\v{y}|\v{x}} \implies g_{\v{x}}^{-1} \cdot \v{y}^* \in \argmax_{\v{y}\in \set{Y}} p\pr{\v{y}|\v{x}}
\end{align}
which concludes the proof.
\end{proof}

\begin{lemma}
\label{lemma:2}
Let $ p\pr{\v{y}\mid \v{x}} = p\pr{g\cdot \v{y}\mid g\cdot \v{x}} $ for all $g\in G$. Then,
\begin{align}
\pr{\argmax_{\v{y}\in \set{Y}} p\pr{\v{y}|g \cdot \v{x}}} = g \cdot \pr{\argmax_{\v{y}\in \set{Y}} p\pr{\v{y}|\v{x}}}
\end{align}
\end{lemma}
\begin{proof}
From the symmetry $p$, we have for all $g \in G$
\begin{align}
p\pr{\v{y}\mid g\cdot \v{x}} &= p\pr{g^{-1}\cdot \v{y}\mid\v{x}} 
\end{align}
Therefore,
\begin{align}
\v{y}^* \in \argmax_{\v{y}\in \set{Y}} p\pr{\v{y}|g\cdot \v{x}} \implies g \cdot \v{y}^* \in \argmax_{\v{y}\in \set{Y}} p\pr{\v{y}|\v{x}}
\end{align}
which concludes the proof.
\end{proof}

We now provide the proof of \cref{th:max}.

\vspace{1ex}
\begin{proof}
We have
\begin{align}
\gv{\phi}\pr{\v{x}} = c\pr{\argmax_{\v{y}\in \set{Y}} p\pr{\v{y}|\v{x}}}
\end{align}
Using Lemma \ref{lemma:1}, we therefore have
\begin{align}
\gv{\phi}\pr{\v{x}} &\in G_{\v{x}}\cdot \pr{\argmax_{\v{y}\in \set{Y}} p\pr{\v{y}|\v{x}}}\\
\gv{\phi}\pr{g \cdot\v{x}} &\in G_{\v{x}}\cdot \pr{\argmax_{\v{y}\in \set{Y}} p\pr{\v{y}|g \cdot \v{x}}}
\end{align}
Using Lemma \ref{lemma:2}, we obtain
\begin{align}
\gv{\phi}\pr{g \cdot\v{x}} &\in g\cdot G_{\v{x}}\cdot \pr{\argmax_{\v{y}\in \set{Y}} p\pr{\v{y}|\v{x}}}
\end{align}
Using the assumption that the $\argmax$ is only one orbit, we have
\begin{align}
\gv{\phi}\pr{g \cdot\v{x}} &\in g\cdot G_{\v{x}}\cdot c\pr{\argmax_{\v{y}\in \set{Y}} p\pr{\v{y}|\v{x}}}\\
\gv{\phi}\pr{g \cdot\v{x}} &\in g\cdot G_{\v{x}}\cdot \gv{\phi}\pr{\v{x}}
\end{align}
This is equivalent to saying that there exists a $g_2 \in g\cdot G_{\v{x}}$ such that
\begin{align}
\gv{\phi}\pr{g \cdot\v{x}} &= g_2 \cdot \gv{\phi}\pr{\v{x}}
\end{align}
which is the relaxed equivariance condition.
\end{proof}

\subsection{\cref{th:mlp}}
\label{proof:mlp}

\begin{proof}

First, we show that if the condition \cref{eq:coset} is satisfied, then for all $g_1 \in G$ and for all $\v{x} \in {\set{X}_{K}}$, there exists a $g_2\in g_1 K$ such that the constraint
\begin{align}
\rho'\pr{g_2} \v{W} \v{x} = \v{W} \rho\pr{g_1} \v{x} \label{eq:constraint}
\end{align}
is satisfied.

For some $g_1\in G$, consider the set of elements that belong to the same coset of the stabilizer of $\v{x}$, e.g. the set $g_1 K$. For all these group members, the constraint \cref{eq:constraint} can be satisfied with the same $g_2$. We can therefore have for all these elements,
\begin{align}
\rho'\pr{g_2} \v{W} \v{x} = \v{W} \rho\pr{g_2 \cdot k} \v{x},
\end{align}
where $k \in K$ and a unique $g_2$ chosen arbitrarily in $g_1 K$.
By definition of the stabilizer, we have
\begin{align}
\rho'\pr{g_2} \v{W} \v{x} &= \v{W} \rho\pr{g_2} \rho\pr{k} \v{x} \\
\rho'\pr{g_2} \v{W} \v{x} &= \v{W} \rho\pr{g_2} \v{x}
\end{align}
Then, we know that by definition the projection $\v{P}_{\set{X}_{K}}$ maps $\mathbb{R}^n$ onto $\set{X}_{K}$. Thus, for any $\v{y} \in \mathbb{R}^n$, we have
\begin{align}
& \rho'\pr{g_2} \v{W} \v{P}_{\set{X}_{K}} \v{y} = \v{W} \rho\pr{g_2} \v{P}_{\set{X}_{K}} \v{y}\\
& \v{W} \v{P}_{\set{X}_{K}} \v{y} = \rho'\pr{g_2} \v{W} \rho\pr{g_2} \v{P}_{\set{X}_{K}} \v{y}\\
& \pr{\v{W} -  \rho'\pr{g_2} \v{W} \rho\pr{g_2}}\v{P}_{\set{X}_{K}} = 0  \label{eq:constraint_2}
\end{align}

Therefore, if for all cosets in $G/G_{\v{x}}$, \cref{eq:constraint_2} is satisfied with an arbitrary representative, \cref{eq:constraint} is satisfied for all $g_1 \in G$ and $\v{x} \in {\set{X}_{K}}$.

Second, we prove that for all orbits, $O \in \set{X}_{\br{H}}/G$ and for any $K\in \br{H}$, there must be a $\v{x} \in \set{X}_K \cap O$.
For any $O$, consider an arbitrary representative $\v{z}\in O$. It must be that $G_{\v{z}} \supseteq H$ for some $H \in \br{H}$. Since $H$ and $K$ are conjugate, there exists a $g\in G$ such that $gHg^{-1} = K$. Since stabilizers of elements in the same orbit are conjugate, we have $G_{g\cdot \v{z}} \supseteq gHg^{-1} = K$. Therefore, $g\cdot \v{z} \in \set{X}_K \cap O$. 

Finally, we invoke the orbit consistency property (\cref{subsec:orbit}) to show that for any orbit $O \in \set{X}_{\br{H}}/G$, since there is an $\v{x} \in \set{X}_K \cap O$, \cref{eq:constraint_2} must be statisfied for any $\v{x} \in O$. Since this is true for any $O$, \cref{eq:constraint_2} also holds for any $\v{x} \in \set{X}_{\br{H}}$. Therefore, the map $\gv{\phi}: \set{X}_{\br{H}} \to \set{Y}, {\v{x}} \mapsto \v{W} \v{x}$ satisfies relaxed equivariance.
\end{proof}

For the coset containing the identity element, the representative can selected as the identity itself, such that there is no constraint. This therefore results in $\abs{G}/\abs{G_{\v{x}}} -1$ constraints. 

Note that contrarily to standard equivariance constraints like in \citep{finzi2021practical}, it does not follow from these constraints that if
\begin{align}
\pr{\v{W}- \rho'\pr{g_1}^T \v{W} \rho\pr{g_1}} \v{P}_{\set{X}_{K}} = 0,\\
\pr{\v{W}- \rho'\pr{g_2}^T \v{W} \rho\pr{g_2}} \v{P}_{\set{X}_{K}} = 0,
\end{align}
a similar constraint is also satisfied for $g_1 \cdot g_2$. It is therefore not possible to straightforwardly reduce the constraints to a set of generators.



\section{Properties of Relaxed Equivariance}\label{apd:relaxed}

\subsection{Equivariant functions}
\label{subsec:eq}
This property is trivially satisfied, but it is still useful to formulate it explicitly.
\begin{proposition}
Let $\gv{\phi}$ be equivariant. Then, $\gv{\phi}$ satisfies relaxed equivariance.
\end{proposition}
\begin{proof}
If $\gv{\phi}$ is equivariant:
\begin{align}
\gv{\phi}\pr{g\cdot \v{x}} = g\cdot \gv{\phi}\pr{\v{x}}
\end{align}
Since $g\in gG_{\v{x}}$, $\gv{\phi}$ satisfies the relaxed equivariance condition.
\end{proof}

\subsection{Composition}
\label{subsec:comp}
\begin{proposition}
Let $\gv{\phi}_1: \set{X} \to \set{Y}$ and $\gv{\phi}_2: \set{Y} \to \set{Z}$ satisfy relaxed equivariance. Then $\gv{\phi}_2 \circ \gv{\phi}_1$ satisfies relaxed equivariance.
\end{proposition}
\begin{proof}
We have
\begin{align}
\gv{\phi}_2\pr{\gv{\phi}_1\pr{g_1\cdot \v{x}}} = \gv{\phi}_2\pr{g_2\cdot \gv{\phi}_1\pr{\v{x}}}
\end{align}
where $g_2\in g_1G_{\v{x}}$. 
Then,
\begin{align}
\gv{\phi}_2\pr{g_2\cdot \gv{\phi}_1\pr{\v{x}}} = g_3\cdot\gv{\phi}_2\pr{ \gv{\phi}_1\pr{\v{x}}}
\end{align}
where $g_3\in g_2G_{\v{x}}$. Since $g_2G_{\v{x}} = g_1G_{\v{x}}$, we have $g_3\in g_1G_{\v{x}}$ and this completes the proof.
\end{proof}

\subsection{Orbit-consistency}
\label{subsec:orbit}
\begin{proposition}
Let $G$ act on $\set{X}$ and $\set{Y}$. Assume that $G$ acts transitively on $\set{X}$, such that $\set{X}$ is a single orbit. For any $\v{x}\in \set{X}$ and $\gv{\phi}:\set{X} \to \set{Y}$, if $\forall g_1 \in G$ there exists a $g_2 \in g_1G_{\v{x}}$ such that
\begin{align}
\gv{\phi}\pr{{g_1} \cdot \v{x}} = {g_2} \cdot \gv{\phi}\pr{\v{x}},\label{eq:rel_x}
\end{align}
then $\phi $ satisfies the relaxed equivariance condition.
\end{proposition}

\begin{proof}
Any $\v{y}\in \set{X} = G \cdot \v{x}$ can be written as $\v{y} = g\cdot \v{x}$ for some $g\in G$.
We therefore have
\begin{align}
\gv{\phi}\pr{{g_1} \cdot \v{y}} = \gv{\phi}\pr{{g_1} \cdot g\cdot\v{x}}.
\end{align}
From \cref{eq:rel_x}, we have
\begin{align}
\gv{\phi}\pr{{g_1} \cdot \v{y}} = {g_1} \cdot g\cdot g_{\v{x}}\cdot\gv{\phi}\pr{\v{x}}, \label{eq:step_rel_x}
\end{align}
for some $g_{\v{x}}\in G_{\v{x}}$.
From \cref{eq:rel_x}, we also know that
\begin{align}
\gv{\phi}\pr{g\cdot\v{x}} = g\cdot g'_{\v{x}}\cdot \gv{\phi}\pr{\v{x}},
\end{align}
for some $g'_{\v{x}}\in G_{\v{x}}$. Therefore,
\begin{align}
{g'_{\v{x}}}^{-1}\cdot g^{-1}\cdot \gv{\phi}\pr{g\cdot\v{x}} = \gv{\phi}\pr{\v{x}}.
\end{align}
Replacing in \ref{eq:step_rel_x}, we obtain
\begin{align}
\gv{\phi}\pr{{g_1} \cdot \v{y}} = {g_1} \cdot g\cdot g_{\v{x}}\cdot{g'_{\v{x}}}^{-1}\cdot g^{-1}\cdot \gv{\phi}\pr{\v{y}}.
\end{align}
Since we have $g_{\v{x}}\cdot{g'_{\v{x}}}^{-1} \in G_{\v{x}}$ and $g\cdot g_{\v{x}}\cdot g^{-1}\in G_{\v{y}} \forall g_{\v{x}}\in G_{\v{x}}$, we have
\begin{align}
\gv{\phi}\pr{{g_1} \cdot \v{y}} = {g_1} \cdot g_{\v{y}} \cdot \gv{\phi}\pr{\v{y}},
\end{align}
where $g_{\v{y}} \in G_y$. This completes the proof.
\end{proof}

\end{document}